# A NotSo Simple Way to Beat Simple Bench

Soham Sane[1,2] & Angus McLean[1]

Collins Aerospace[1]

*An RTX Business*

## ABSTRACT

This paper presents a novel framework for enhancing reasoning capabilities in large language models (LLMs) by leveraging iterative reasoning and feedback-driven methodologies. Building on the limitations identified in the SimpleBench benchmark, a dataset designed to evaluate logical coherence and real-world reasoning, we propose a multi-step prompting strategy coupled with global consistency checks to improve model accuracy and robustness. Through comparative analysis of state-of-the-art models, including Claude 3 Opus, Claude 3.5, GPT-4o, and o1-preview, we demonstrate that iterative reasoning significantly enhances model performance, with improvements observed in both standard accuracy metrics (AVG@5) and a newly introduced metric, Extreme Averaging (EAG@5). Our results reveal model-specific strengths: Claude excels in maintaining logical consistency, while GPT-4o exhibits exploratory creativity but struggles with ambiguous prompts. By analyzing case studies and identifying gaps in spatial and temporal reasoning, we highlight areas for further refinement. The findings underscore the potential of structured reasoning frameworks to address inherent model limitations, irrespective of pretraining methodologies. This study lays the groundwork for integrating dynamic feedback mechanisms, adaptive restart strategies, and diverse evaluation metrics to advance LLM reasoning capabilities across complex and multi-domain problem spaces.

*1. Affiliation, 2. Lead Author*

# 1  INTRODUCTION

*1.1 – Background*

Large Language Models (LLMs) have demonstrated significant potential in solving complex problems across various domains, from natural language understanding to reasoning and decision-making tasks. Despite their impressive performance, these models often face limitations in maintaining logical coherence and navigating multi-step problem-solving scenarios. Traditional approaches to reasoning in LLMs, such as direct inference or chain-of-thought (CoT) prompting, have provided partial solutions. However, these methods frequently struggle with ensuring global consistency and adapting to ambiguous or novel problems.

The emergence of evaluation benchmarks such as SimpleBench (Phillip & Hemang 2024) has further illuminated these shortcomings. SimpleBench is designed to assess LLMs on their ability to reason through logical tasks, combining real-world contexts with abstract problem-solving. It combines diverse question formats, ranging from spatial reasoning to ethical decision-making, to test a model's ability to navigate nuanced problem spaces. Unlike benchmarks focused solely on accuracy, SimpleBench emphasizes the importance of logical coherence and contextual understanding. While current frontier models have shown promise, their reasoning often falters in edge cases where nuanced understanding and iterative refinement are crucial. These observations underscore the need for enhanced frameworks capable of overcoming these barriers.

*1.2 Problem Statement*

Existing methods for reasoning in LLMs rely heavily on static or minimally iterative frameworks, which lack the flexibility to refine solutions dynamically. This rigidity limits their ability to address complex reasoning tasks, particularly those requiring exploration of alternative assumptions or iterative corrections. Additionally, current benchmarks, including SimpleBench, reveal that while models may perform well under standard metrics, their performance in scenarios demanding robustness and adaptability remains suboptimal.

To address these challenges, we propose an iterative reasoning framework that introduces a structured process of multi-step prompting, feedback validation, and global consistency checks. By leveraging these mechanisms, our approach seeks to enhance the logical coherence, adaptability, and overall robustness of LLMs in tackling complex reasoning tasks.

## *1.3 Related Work*

Significant research efforts have been directed toward improving LLM reasoning capabilities. Chain-of-Thought (CoT) prompting, as introduced by (Wei et al. 2022), allows models to break down problems into sequential reasoning steps, improving performance on arithmetic and commonsense reasoning tasks. However, CoT relies on static reasoning chains that cannot adapt dynamically during inference, leading to challenges in handling ambiguous or complex inputs.

A 2024 study also attempted iterative CoT prompting with integrated feedback loops to refine reasoning during inference (Sun et al. 2024). While this approach introduced dynamic correction capabilities, it was limited in scalability and often suffered from inconsistencies in complex multi-step tasks. Our framework diverges by introducing a dedicated external module for iterative refinement, enabling more robust and globally consistent reasoning outcomes.

Our methodology leverages multi-step prompting, feedback mechanisms, and global consistency checks to iteratively enhance reasoning quality. This dynamic adaptability positions our approach as a novel contribution to LLM reasoning frameworks.

## *1.4 Purpose and Goals*

This paper aims to address the limitations of existing reasoning methodologies by introducing an iterative reasoning framework. The primary goals are to enhance the logical coherence, adaptability, and robustness of LLMs and to provide a structured mechanism for handling complex reasoning tasks. Through empirical evaluation using SimpleBench, we aim to demonstrate the effectiveness of our approach and establish a foundation for future advancements in LLM reasoning capabilities.

---

*The rest of this page is left intentionally blank*

# 2 SOLUTION OVERVIEW

To account for the lack of logic and reasoning in the current space of LLM infrastructures, we decided to choose an approach that closely mirrors and mimics how a human would respond to a given prompt. This involved the use of multi-iterative prompting and feedback analysis through step-by-step reasoning.

## 2.1 – Model Infrastructure

The solution implemented a multi-step reasoning process using a baseline model, GPT-4o & Claude 3 Opus, to query a given prompt. The baseline models were chosen as they were the last iteration of models by OpenAI & Anthropic that were not trained on a system of reasoning steps. Unlike conventional one-shot inference methods, where the model directly predicts an answer, this approach prompts the model to generate reasoning steps iteratively. Each reasoning step builds upon the previous steps, ensuring a gradual and logically consistent derivation of the solution. This approach leverages the strengths of chain-of-thought (CoT) reasoning, enabling the model to break down complex problems into manageable components.

The architecture consists of several interconnected modules: the step generation module, the feedback gate, the global consistency check mechanism, and the final solution derivation component. A restart counter and step limiter enforce control over computational resources and iterative depth. These modules collectively facilitate a structured, iterative problem-solving framework. The activity diagram, found in section 7.1, can be viewed in the appendix of this paper.

### 2.1.1 – Step Generation

Step generation is the foundational component of the model infrastructure. For each problem, the model is prompted to generate reasoning steps sequentially, starting from the first logical step. If previous steps and their outputs exist, they are passed to the model as context, allowing the generation process to build incrementally. Each step is formulated using a structured prompt, explicitly instructing the model to consider environmental, contextual, and real-world factors that may influence the solution.

The model outputs reasoning in the format: "Step X: [Your reasoning here]." This ensures clarity and logical progression. The step generation module also incorporates mechanisms to detect when no further steps are required, using an early termination signal, NO_MORE_STEPS, to avoid unnecessary computations.

### 2.1.2 – Feedback Gate

The feedback gate serves as a validation mechanism for the reasoning steps. Each reasoning step is evaluated against the problem context and previous steps to ensure logical consistency and adherence to natural laws. This gate identifies flaws, incorrect assumptions, or incomplete reasoning within each step.

The feedback process uses a structured prompt that evaluates the latest step against specific criteria, such as:

1. Logical consistency with prior steps.
2. Alignment with the problem's constraints and context.
3. Adherence to physical laws and reasonable assumptions.

If the feedback gate identifies issues, the model generates a revised step based on the feedback provided. This iterative correction process ensures that flawed reasoning is corrected dynamically within the chain itself, maintaining the trail of thought even if the solution is only partially accurate.

### 2.1.3 – Global Consistency Check

The global consistency check consolidates and evaluates all reasoning chains generated during the problem-solving process. This module examines the chains for discrepancies, unaddressed assumptions, or unexplored logical paths. The evaluation process uses a structured prompt to:

1. Identify incorrect or unstated assumptions.
2. Compare reasoning chains for consistency and logical coherence.
3. Propose alternative focuses or restarts, if necessary.

The global consistency check ensures that the solution space is explored, and any unexplored assumptions & possible deviations are addressed. If no further assumptions remain, the module synthesizes a final answer by integrating the most robust reasoning chain.

### 2.1.4 – Final Solution Derivation

Once all reasoning chains have been evaluated, the final solution derivation module selects the most logical and consistent chain as the final answer. This selection process prioritizes solutions that demonstrate completeness, alignment with the problem context, and logical soundness through a scoring mechanism and careful specific prompting. This involves the comparison of solution chains that may disprove other solution chains despite scoring higher in a quantitative metric.

### 2.1.5 – Restart Counter & # Steps Limiter

To manage computational resources effectively, the model infrastructure incorporates a restart counter and steps limiter. The restart counter allows the reasoning process to reset and explore alternative assumptions or logical paths. The steps limiter restricts the depth of iterative reasoning within each restart, preventing unnecessary computational overhead.

By enforcing these limits, the system ensures a balance between thoroughness and efficiency. The restart mechanism integrates identified assumptions and their inverses into subsequent iterations, progressively refining the solution space.

## 2.2 – Prompting Methods

The prompting methods employed in the system are designed to maximize clarity and logical progression. Structured prompts explicitly guide the model's reasoning process, ensuring that each step adheres to the desired format and context. Prompts are customized for different modules via their temperature and presence penalty settings, such as step generation, feedback validation, and global consistency checks, aligning with their specific objectives.

For example, the step generation prompts instruct the model to consider assumptions, constraints, and real-world factors, while the feedback gate prompts focus on validating logical consistency and identifying flaws. These tailored prompts play a crucial role in maintaining the coherence and reliability of the reasoning process. Specific details about the prompts are available in the appendix of this paper.

---

*The rest of this page is left intentionally blank*

# 3 FINDINGS & ANALYSIS

### 3.1 – Scoring

To evaluate the performance of the proposed iterative reasoning framework, we employed two distinct scoring methods: **AVG@5** & **Rounded EAG@5**. These metrics were selected to capture different aspects of model performance and provide robust insights, especially given the constraints of limited trial data.

#### 3.1.1 - AVG@5

AVG@5 calculates the average performance score over five independent trials, offering a straightforward measure of consistency across repeated iterations. This metric provides a baseline for comparing model reliability and is particularly useful when data availability is limited.

#### 3.1.2 - Rounded EAG@5

Rounded EAG@5 (Extreme Averaging) was introduced as a novel metric to focus on the extremes of model performance. Unlike AVG@5, which averages across all trials without weighting, EAG@5 adjusts scores based on their relative extremity. This metric emphasizes scenarios where models perform either exceptionally well or consistently poorly, rewarding stability while penalizing failures to meet minimum thresholds.

The scoring formula for Rounded EAG@5 is defined as:

$$\boldsymbol{EAG@5} = \begin{cases} -0.25 \text{ if } AVG = 0, \\ 0.5 * AVG \text{ if } 0.33 \leq AVG, \\ 0.75 * AVG \text{ if } 0.33 < AVG \leq 0.66, \\ 1.5 * AVG \text{ if } AVG > 0.66 \end{cases}$$

This design rewards models for being consistently good across trials while introducing penalties for failing to achieve any success. By focusing on extreme cases, Rounded EAG@5 captures performance variability that may be missed by simpler averaging methods.

#### 3.1.3 - Rationale for Metric Selection

The decision to use AVG@5 and Rounded EAG@5 was driven by several practical and methodological considerations:

1. Limited Trial Data: As this study was self-funded, extensive trial runs were not feasible. Rounded EAG@5 offers a more granular view of

performance under such constraints, particularly for models that may excel in edge cases or fail sporadically.
2. Avoidance of MAG (Majority Averaging): While MAG has been previously used in the SimpleBench study, our analysis of its results indicated consistent alignment with AVG. Thus, AVG@5 was deemed sufficient for capturing general trends without the added complexity of MAG.
3. New Insights through Extremes: Rounded EAG@5 was developed as an exploratory metric to potentially reveal insights that AVG or MAG might overlook. By amplifying the impact of extreme outcomes, it provides a nuanced perspective on model stability and robustness.

### 3.2 – Models of Comparison

Four large language models were selected for comparison in this study: **o1-preview**, **Claude 3.5**, **Claude 3 Opus**, and **GPT-4o**. The selection aimed to examine the differences in performance between models with and without explicit Chain-of-Thought (CoT) enhancements and to evaluate the effectiveness of the proposed prompting methodology on baseline and CoT-enabled models. It can be theorized that 4o & Claude 3 Opus were not explicitly trained to synthesize reasoning steps and therefore were chosen as the baseline models for the application of our solution. The models that were used as a baseline were prompted using techniques consistent with those employed in the SimpleBench study.

#### 3.2.1 – Objective of Comparison

The primary objective was to evaluate how iterative prompting strategies could enhance the performance of baseline models. Essentially, can structured prompting and iterative refinement elevate the reasoning capabilities of baseline models to levels comparable with CoT-enhanced models? Can our prompting solution act as a meta-layer of reasoning enhancement, decoupled from the underlying model architecture?

## 3.3 - Results

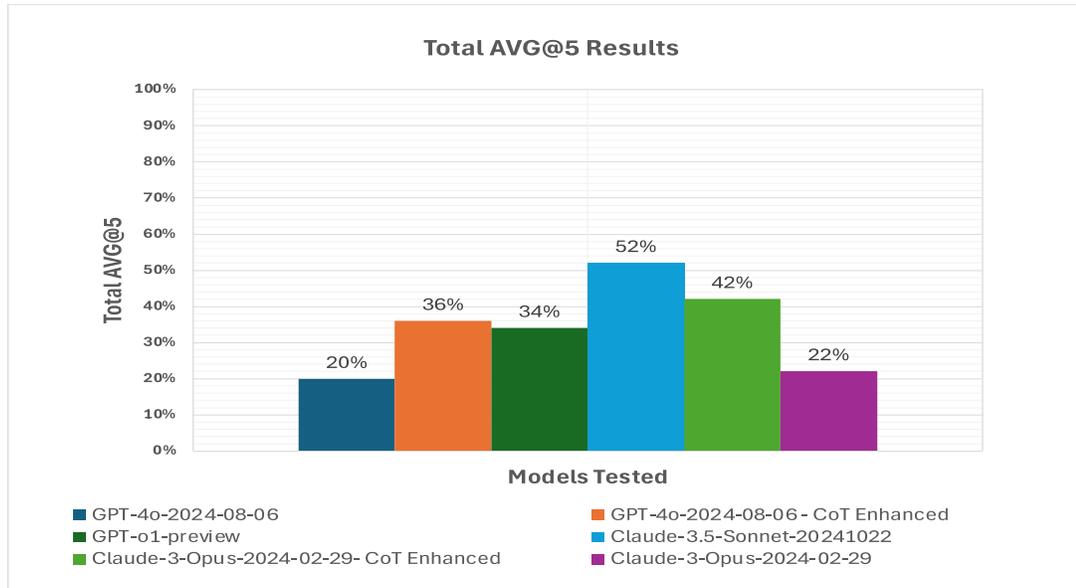

*Figure 1: Total AVG@5 Results for Models Tested*

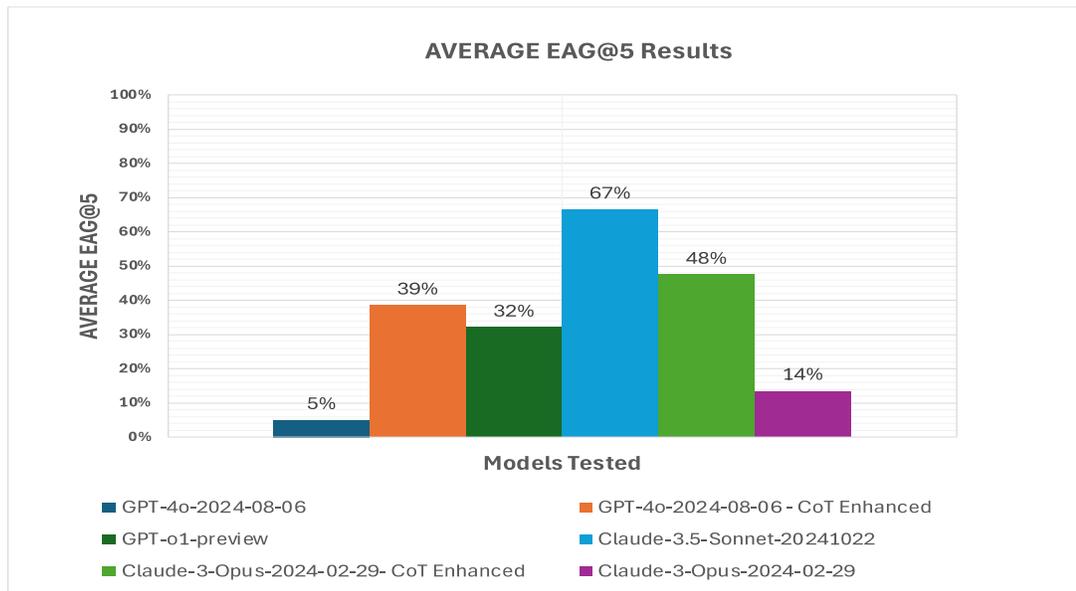

*Figure 2: Average EAG@5 Results for Models Tested*

The bar graphs, figure 1 & figure 2, highlight the results found across testing at 5 trials on the public dataset on SimpleBench. It is interesting to note that the CoT Enhanced models (our solution) consistently scored similar to their next generation counterparts. It is also important to remember that this data set only consists of 50 total trials per graph and therefore has high variability

### 3.3 – Findings

The findings from this study reveal distinct patterns in model behavior and performance, offering valuable insights into their reasoning capabilities and areas of divergence.

#### 3.3.1 – o1-preview Performance vs. GPT-4o

The o1-preview model demonstrated notable performance improvements over GPT-4o, potentially indicating that o1 builds upon GPT-4o as a foundational architecture while incorporating reasoning chains identified during inference. This observation raises an important question: could scaling baseline models like GPT-4o result in comparable enhancements to reasoning capabilities?

Another intriguing aspect is whether o1's reasoning relies on an iterative process that utilizes intermediate outputs to inform subsequent steps or if it primarily employs a one-shot approach. Insights from OpenAI's playground, which tracks unique reasoning tokens for outputs but not inputs, suggest a predominance of one-shot reasoning (Brown et al., 2020). In contrast, the iterative framework examined in this study offers a compelling alternative. Although computationally more demanding, such a framework could achieve superior accuracy with adequate training resources. Moreover, iterative feedback loops might address the creativity limitations identified in recent comparisons between o1 and GPT-4o, further enhancing o1's capacity for diverse reasoning tasks ((Li et al., 2024).

#### 3.3.2 - Behavioral Divergences Between GPT-4o and Claude

The user-observed analysis of GPT-4o and Claude models, when enhanced, also revealed a striking divergence in their reasoning approaches:

- **Exploratory Reasoning in GPT-4o:**
  GPT-4o exhibited a more creative and exploratory reasoning style, often attempting to contextualize the problem with broader assumptions. This approach reflects its design, which prioritizes flexibility and a wide-ranging understanding of prompts. While this creativity proved advantageous in ambiguous scenarios, it occasionally led to overcomplications in structured benchmarks, where focused reasoning was required.
- **Objective and Consistent Claude:**
  In contrast, Claude models demonstrated a more objective and structured reasoning approach. Their outputs were marked by consistent logical alignment, with minimal deviations or restarts. This reliability reflects their optimization for precision and stability, likely a result of their training framework emphasizing adherence to defined reasoning paths. However, this confidence sometimes limited their ability to explore alternative solutions, which could be a disadvantage in more open-ended or creative tasks.

The differences in reasoning styles underscore a potential opportunity for complementary use. For instance, GPT-4o's exploratory behavior might compensate for Claude's tendency toward rigidity, while Claude's consistency could ground GPT-4o's more expansive reasoning. Their use in tandem could offer an even more lucrative score on SimpleBench.

### 3.3.3 - Potential Impacts of Restarts and Exploratory Reasoning

The findings also highlight a potential utility of strategic restarts during reasoning. In certain test cases, models like GPT-4o and Claude reached a near-perfect score (4/5) but failed to achieve full marks due to a unique but inaccurate initial step generation caused by the high temperature setting. However, forcing restarts during early reasoning stages, particularly with higher temperature values, may foster the exploration of alternative solutions, increasing the likelihood of identifying optimal reasoning chains.

### 3.4 – Specific Question Analysis

To better understand how different models perform on complex reasoning tasks, we selected two prompts from our dataset as representative examples. These prompts explore various aspects of reasoning, including the ability to interpret situational nuances and safety-critical decision-making.

### Question 9-

The prompt for Question 9 presented a logistical scenario:

*Agatha makes a stack of 5 cold, fresh single-slice ham sandwiches (with no sauces or condiments) in Room A, then immediately uses duct tape to stick the top surface of the uppermost sandwich to the bottom of her walking stick. She then walks to Room B, with her walking stick, so how many whole sandwiches are there now, in each room?*

A. 4 whole sandwiches in Room A, 0 whole sandwiches in Room B
B. No sandwiches anywhere
C. 4 whole sandwiches in Room B, 1 whole sandwich in Room A
D. All 5 whole sandwiches in Room B
E. 4 whole sandwiches in Room B, 1 whole sandwich in Room A
F. All 5 whole sandwiches in Room A

**Answer**: *A*

When reasoning through this prompt, the OpenAI family of models attempted to extract more context about the situation, such as whether sandwiches adhered to specific conditions (e.g., freshness or whether duct tape could functionally hold them together during movement). This need for context could reflect an opportunity for improvement, suggesting that models might benefit from explicitly querying users during reasoning. Allowing models to ask clarifying questions could make their reasoning steps more robust and user-aligned, especially in scenarios where additional situational details are relevant.

On the contrary, the Claude family of models understood the situation in the sense of an object was being taped yet failed to even suggest variations on how that taping might alter the final solution as seen by its objective tone. It is also interesting to note that OpenAI models were able to identify the inconsistency in the answer choices, but Claude was not.

*Question 5-*
The prompt for Question 5 explored moral and safety reasoning:

*Peter needs CPR from his best friend Paul, the only person around. However, Paul's last text exchange with Peter was about the verbal attack Paul made on Peter as a child over his overly-expensive Pokémon collection, and Paul stores all his texts in the cloud, permanently. Paul will [ _ ] help Peter.*

A. Probably not
B. Definitely
C. Half-heartedly
D. Not
E. Pretend to
F. Ponder deeply over whether to

**Answer**: *B*

The Claude family of models excelled in this question, consistently prioritizing safety and ethical responsibility in its reasoning. o1-preview, while adequate, displayed variability in its step-by-step reasoning, with occasional deviations into tangential considerations about the emotional context. GPT-4o, on the other hand, offered a middle-ground response, occasionally overanalyzing irrelevant details.

This variation in responses raises questions about how red-teaming and training biases influence model behavior. For example:

- Claude's performance suggests its training emphasized safety-critical outcomes, possibly reflecting an overcorrection from adversarial testing aimed at avoiding harmful outputs. However, such emphasis might limit its exploratory reasoning capabilities.
- o1-preview's deviations suggest it was designed to accommodate diverse perspectives, but this flexibility might come at the cost of consistent safety-focused reasoning. However, it is interesting to note that our CoT solution on 4o still performed at the same level as o1 which was higher than the baseline 4o model.

This raises an important question: Were these models trained with a bias toward positivity and safety, focusing on producing safe outputs, or were they designed to filter potentially harmful prompts before reasoning begins? This distinction is critical for understanding how models manage exploratory deviations versus enforcing structured, outcome-driven paths. Moreover, how can reasoning step generation ensure safe exploration of diverse solution sets without inadvertently entertaining harmful or logically flawed chains?

### 3.5 – Partial Credit

One intriguing aspect of the study emerged when analyzing cases where a model provided strong reasoning or feedback but ultimately arrived at an incorrect solution. This raises the question of how performance should be defined to account for the "work" and not only the final solution.

#### *Defining Performance Without Predefined Choices-*

In our evaluation settings, performance is binary—either correct or incorrect. However, in reasoning-based tasks where models must navigate complex problem spaces, this strict dichotomy may overlook nuanced progress toward the correct answer. For instance, a model might produce reasoning chains that, while not entirely correct, offer substantial hints or partial solutions that align with the correct outcome.

***Enhancing Situational Awareness Through Queries-***
The observations from the study also highlight a potential gap in how models address uncertainties. For example, instead of committing to a potentially flawed reasoning chain, a model could be trained to proactively query users about ambiguities or uncertainties it encounters. This approach would align reasoning processes more closely with human problem-solving behaviors, fostering greater situational awareness. For instance, in cases where contextual clarity is lacking, a model could suggest multiple potential solutions or pose clarifying questions to refine its understanding of the problem as a form of active RLHF.

Such an approach not only enhances the model's reasoning capacity but also engages the user as an active participant in the problem-solving process. Future studies could explore methodologies for training models to incorporate uncertainty handling and dynamic querying as integral components of their reasoning frameworks.

## 3.6 – Future Studies
These findings open several promising avenues for further exploration and improvement of reasoning models. Future work could focus on diversifying the models and techniques used, refining evaluation metrics, and expanding the scope of testing to enhance our understanding of how reasoning frameworks can be optimized.

### 3.6.1 - Exploring Alternative Baselines and Hybrid Approaches
A key area for future research involves a mix-and-match approach, where Chain-of-Thought (CoT) based models are employed selectively—for example, using one model for reasoning steps and another for feedback or global consistency checks. This "CoT within a CoT" strategy could leverage the unique strengths of different models. Furthermore, simply even integrating GPT and Claude into a single framework, as previously discussed, could reveal interesting solutions.

### 3.6.2 - Experimentation with Parameters and Metrics
Another avenue involves systematically testing different p-values and temperature settings, particularly in relation to the holistic feedback gate and initial step generation (Mehta et al., 2023). Fine-tuning these parameters could uncover the ideal configurations for fostering diverse yet coherent reasoning paths. Expanding the scope of the SimpleBench study to include more questions and additional models would also provide a richer dataset for comparison.

Given the limitations in funding and model access, future efforts should consider better data utilization through metrics such as MAG (Median Aggregated Grade). Testing MAG across answer choices, instead of limiting it to Boolean evaluations, could offer deeper insights into how consistently reasoning paths align with correct solutions. This analysis could also introduce a Creativity Index to evaluate models' capacity to generate innovative or unexpected reasoning steps.

### *3.6.3 - Implications for Task-Specific Agent Development*

Finally, this study underscores the potential of CoT prompting, even on baseline models, when tailored to specific user needs. By refining reasoning frameworks for task-specific applications, we could significantly improve the efficiency and accuracy of agent-based systems. This adaptability, especially using output feedback to guide next step generation, may be critical for deploying AI agents in specialized roles where precision and contextual understanding are paramount.

---

*The rest of this page is left intentionally blank*

# 4 IMPROVEMENTS

### *4.1 - Specifically Trained Models*

One possible improvement that can be made in the current implementation of our system is to use specifically trained models for step generation (Cheng et al., 2024). These models can be trained not only to generate intermediate reasoning steps but also to derive feedback and comprehend the assumptions underlying their outputs. In simpler terms, having a contextually aware and specifically trained model on how to generate reasoning steps that mimic Chain-of-Thoughts may provide more accurate solutions and pathways. This strategy combined with the recent studies in Chain-of-Thought (CoT) prompting may show that such step-by-step generation could significantly improve task performance in language models from an end-to-end user perspective ([Wei et al., 2022](#)).

Another novel concept involves the use of a dedicated end-gate model that synthesizes the outputs of multiple CoT's to produce a coherent and validated final solution. Similar to the global consistency check that is used in our solution, we can provide this end-gate model with a diverse set of reasoning paths to ensure consistency, robustness, and an improved capacity for error detection. This end gate model falls in the same family as the specifically trained models that are small at scale yet domain rich in their specific role.

This inherently brings forward the question of the Chain-of-Thought process itself. Why is it that we are judging the performance of this model based on the final solution that the CoT produced? A question that arises in our discussions is what if we rather trained our specific models to synthesize multiple CoT's that reach the same yet correct final solution? Instead of treating loss as the distance from the final solution, we should rather train the loss on the quality of the CoT's it produces to reach that final solution. This loss term however is just a fractional representation of all the specific models involved in producing the final result. The division of this loss term would need to be understood as a combined value between the accuracy of the step generated, the logical flow of all steps, and the effectiveness of the global consistency check. This self-learning paradigm could bridge the gap between pre-defined human heuristics and emergent model-driven logic.

### 4.2 - Incorporating Assumptions, Consistency, and Diversions into Feedback Loops

Efficiently handling assumptions, inconsistencies, and diversions within the reasoning process is critical for optimizing the CoT system (Zhang et al., 2023). Our current method involves restarting the CoT generation process whenever a logical inconsistency or error is identified. However, this approach is computationally expensive and may lead to redundant evaluations.

We propose a more efficient strategy where the feedback mechanism has the authority to generate alternative CoTs at the specific point of divergence, rather than restarting the entire process. By delegating the authority to address diversions to the feedback gate, the model can target critical junctures without reprocessing the entire reasoning chain. This method reduces computational overhead while maintaining high solution fidelity as CoT's are generated recursively as they are identified.

Furthermore, by restricting global evaluations to the end-gates of CoTs (start node – end node path), the system can streamline its validation process, focusing only on how the synthesized outputs differ from where they diverged in their reasoning steps. A strong representation of this can be seen in figure 3, where a heat map represents the abstract solution space in which solutions might exists. The "SN" nodes represent the steps we take to reach the final nodes, denoted by "FN".

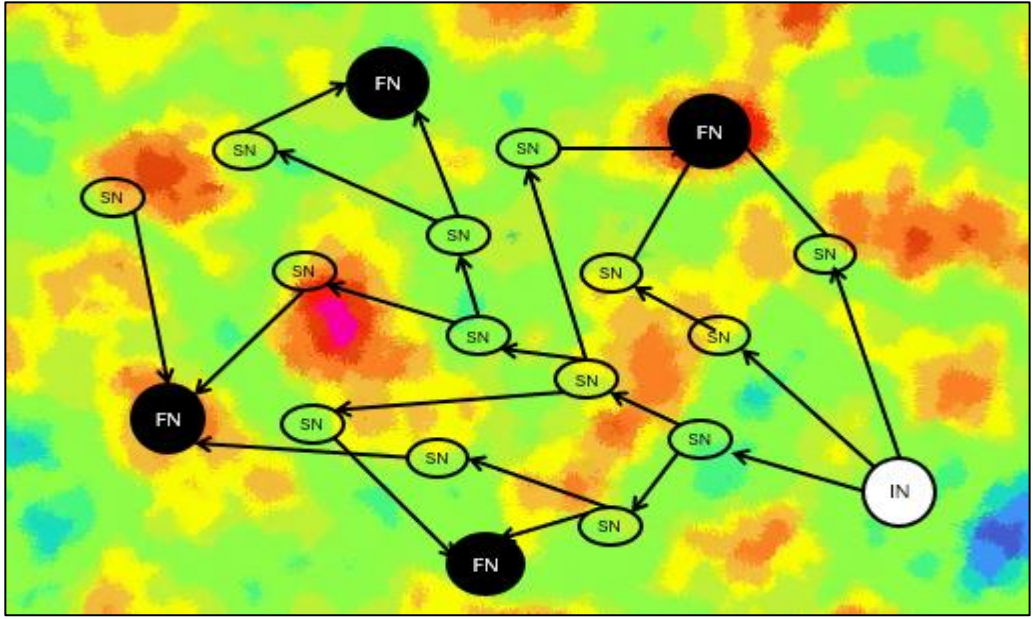

**Figure 3 - Abstract Representation of Solution Space**

Another technique involves utilizing Directed Acyclic Graphs (DAGs) to represent dependencies between reasoning steps. By mapping tasks as nodes and their relationships as edges, we can identify independent steps that can be processed in parallel. This approach ensures logical consistency while significantly reducing execution time for independent tasks.

Parallel decomposition methods also provide avenues for distributing workload. For instance, tasks that exhibit temporal or logical independence can be grouped and executed simultaneously, streamlining multi-step problem-solving. By leveraging parallel execution, the system can process multiple reasoning paths concurrently and reconcile their outputs in subsequent stages.

Additionally, we propose a novel method inspired by hierarchical abstraction: identifying "chunks" of a reasoning branch that can be preprocessed at higher levels of granularity before delving into finer details. This technique mirrors the process of outlining an idea before filling in intricate details—it allows the model to validate the broad structure of reasoning early on, ensuring alignment before investing computational resources into granular evaluations. Such an approach not only improves efficiency but also enhances the overall coherence of the reasoning process by catching structural errors early.

### 4.3 - Context Expansion & Situational Awareness

One of the hallmarks of human reasoning is the ability to dynamically expand the context of a question by recalling relevant knowledge. For instance, when solving a physics problem, individuals instinctively draw upon applicable physical laws and prior experiences. Mimicking this cognitive process in AI could significantly enhance its problem-solving capabilities.

We propose the development of another specifically trained 'system' trained to expand the context of a given prompt as a preliminary step (Diao et al., 2024). This system would generate multiple potential assumptions by first broadening the scope of relevant information. Such a mechanism could increase the richness and depth of reasoning pathways from the start of a solution derivation, enabling the system to address complex and ambiguous queries more effectively.

This context-expansion capability aligns with the notion of attention mechanisms in transformers, where selective focus on relevant inputs has already demonstrated success in natural language understanding ([Vaswani et al., 2017](#)). Expanding more on that same concept could involve enriching attention via querying the correct situational context given an input prompt. Training models specifically to emulate human-like context expansion could further refine their ability to adapt to diverse and inconclusive problem spaces.

### 4.4 – Human Baselines for CoT's

The goal of our study was not truly to conclude on the final solution but rather raise awareness on the possibility of other solutions as the human would have. This notion is key in improving future models through the methods proposed above by focusing on mimicking logical reasoning in a system sense rather than focusing on the outcome.

An interesting thought however is the assumption that human CoT, through which most of the consistency check mimics, is the strongest baseline. As discussed previously, allowing models to independently learn and optimize their reasoning pathways could surpass human-designed heuristics in certain contexts. This goes hand in hand with the observation that we should rather focuses on deriving multiple pathways reaching a correct solution rather than the perfection of a solution. This approach emphasizes the importance of providing models with the flexibility to generate and validate diverse reasoning chains, rather than strictly adhering to human-established patterns.

### 4.5 - Cost and Scaling

It is no secret that the computational and financial costs associated with advanced CoT models present significant challenges. Training models to perform iterative reasoning and context expansion demands substantial resources, necessitating strategies to improve efficiency. For example, figures 4 & 5 show an abstracted graph of computational cost of the three systems described in this paper as the context increases:

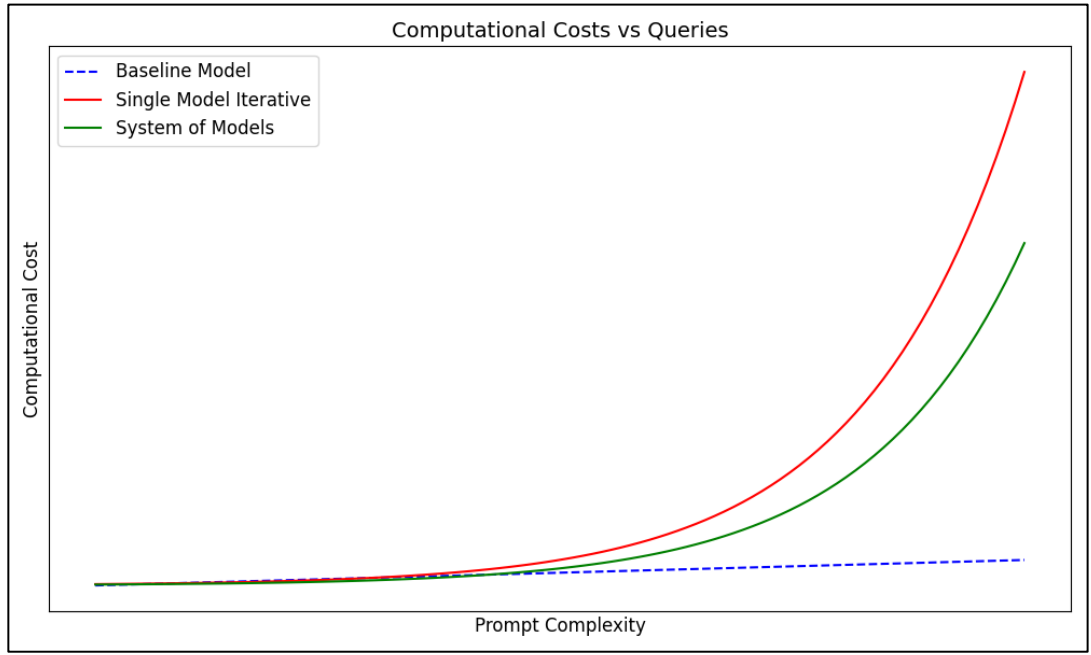

Figure 4 – Computational Costs vs Prompt Complexity

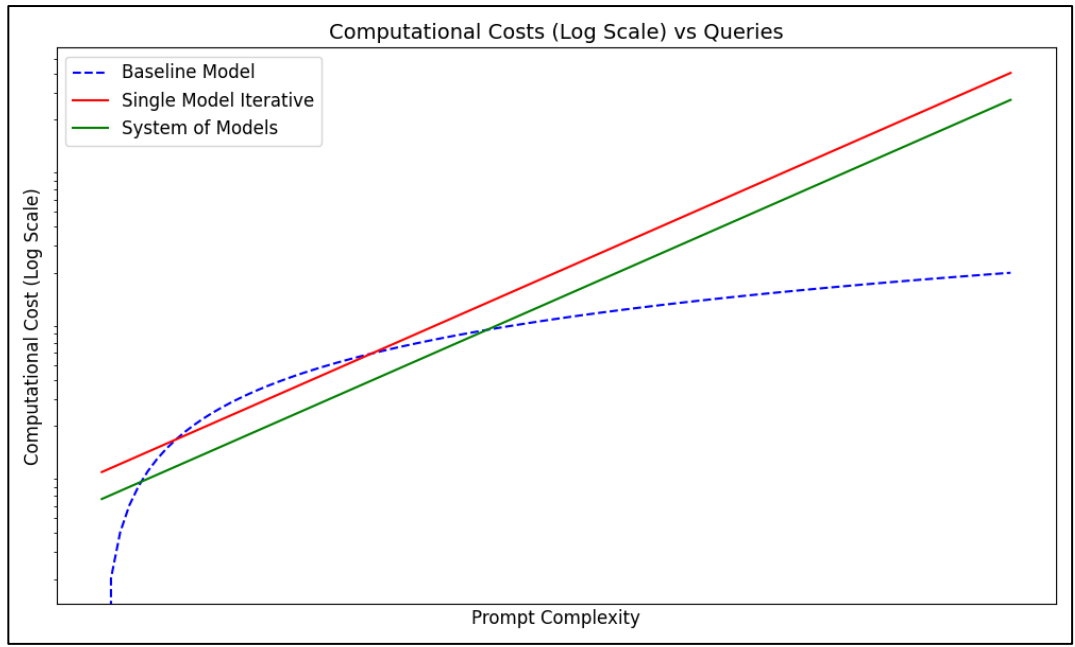

Figure 5 – Computation Costs (Log Scale) vs Prompt Complexity

Scaling remains a double-edged sword in the pursuit of higher intelligence. While larger models excel in objective tasks like mathematics, their performance on reasoning tasks often lags human baselines. This discrepancy suggests that scaling alone may not suffice to bridge the gap. Instead, targeted scaling—focused on enhancing reasoning and evaluation capacities—could unlock new levels of performance without requiring an exponential increase in general training datasets. The development of reasoning-specific datasets and architectures may hold the key to advancing AI's capabilities in this domain ([Bommasani et al., 2021](#)). While this may be true for the specific models in our system, it is still important that we continue to scale the underlying baseline model with as much data through as many domains of knowledge as possible.

---

*The rest of this page is left intentionally blank*

# 5  CONCLUSIONS

**5.1 – Role of LLMs**
The role of Large Language Models (LLMs) in advancing artificial general intelligence (AGI) is multifaceted, particularly when evaluating their utility as standalone transformers versus their potential as integrated systems. LLMs have the potential to function not only as models for general-purpose tasks but also as dynamic systems capable of leveraging domain-specific knowledge and contextual understanding to perform more sophisticated reasoning (Brown et al., 2020; Kaplan et al., 2020).

**5.2 – LLMs as Systems**
LLMs can be viewed in two distinct paradigms: as general transformers with broad but shallow applicability, and as systems tailored to integrate domain-specific knowledge, which can significantly enhance their performance. This distinction underscores the importance of prompt engineering. When a human prompter possesses domain-specific expertise, they inherently guide the model to produce more accurate reasoning paths or Chain-of-Thoughts (CoTs). This suggests that the next logical step in advancing toward AGI may involve enabling models to autonomously generate and differentiate CoTs with greater precision.

By understanding LLMs as **systems** rather than isolated entities, their potential can be further unlocked. This systems-based approach could involve granting models access to external sources, such as the internet, codebases, or historical conversations as we have seen with the recent developments by OpenAI (Komeili et al., 2021; Lazaridou et al., 2022). These capabilities allow LLMs to dynamically retrieve relevant information, facilitating a deeper situational awareness (Guu et al., 2020; Lewis et al., 2020). Combining this ability with the CoT system as described in this paper, we may possess the key to unlocking the principles of AGI, where the system not only answers queries but actively builds and refines its knowledge base to address increasingly complex scenarios.

**5.3 – Context Expansion and AGI Scaling**
Scaling towards AGI is often discussed in terms of increasing the model's parameter count along with increasing its dataset token count. However, this approach overlooks an equally critical aspect: scaling contextual awareness. Models that can store, recall, and efficiently manage context over extended interactions are likely to outperform models focused solely on parameter expansion. Context expansion could involve maintaining memory of prior CoTs, solutions, and even user-specific prompting patterns (Rae et al., 2020).

Such memory capabilities would enable models to revisit prior assumptions, identify overlooked areas, and iteratively refine solutions. For instance, a model that understands a user's habitual framing of questions within a specific domain can preemptively explore assumptions and offer more personalized solutions. This ability to adaptively expand upon past CoTs provides a pathway for models to break free from repetitive loops, particularly in iterative user interactions.

Improvements in contextual awareness have consistently shown direct gains in performance (Lample et al., 2019; Guu et al., 2020). For example, research on retrieval-augmented models ([Guu et al., 2020](#)) highlights how access to relevant context enhances reasoning capabilities. Similarly, memory-augmented transformers ([Rae et al., 2020](#)) demonstrate that storing and recalling extended sequences contributes to more coherent and informed outputs.

### 5.4 – Final Thoughts

The journey toward AGI involves more than scaling parameters—it requires reimagining how models interact with context and domain knowledge. By treating LLMs as systems with dynamic access to external sources and equipping them with advanced context-handling capabilities, we move closer to models capable of human-like reasoning. This approach emphasizes that AGI is not merely about making models larger but about making them smarter, more adaptable, and deeply integrated into their operational environments.

---

*The rest of this page is left intentionally blank*

# 6 REFERENCES

Philip, & Hemang. (2024). SimpleBench: The Text Benchmark in which Unspecialized Human Performance Exceeds that of Current Frontier Models. Retrieved from https://simple-bench.com/.

Brown, T., Mann, B., Ryder, N., Subbiah, M., Kaplan, J., Dhariwal, P., Neelakantan, A., Shyam, P., Sastry, G., Askell, A., Agarwal, S., Herbert-Voss, A., Krueger, G., Henighan, T., Child, R., Ramesh, A., Ziegler, D., Wu, J., Winter, C., Hesse, C., Chen, M., Sigler, E., Litwin, M., Gray, S., Chess, B., Clark, J., Berner, C., McCandlish, S., Radford, A., Sutskever, I., & Amodei, D. (2020). Language Models are Few-Shot Learners. Advances in Neural Information Processing Systems, 33, 1877-1901.

Kaplan, J., McCandlish, S., Henighan, T., Brown, T. B., Chess, B., Child, R., Gray, S., Radford, A., Wu, J., & Amodei, D. (2020). Scaling Laws for Neural Language Models. arXiv preprint arXiv:2001.08361.

Wei, J., Wang, X., Schuurmans, D., Bosma, M., Ichter, B., Xia, F., Chi, E., Le, Q., & Zhou, D. (2022). Chain of Thought Prompting Elicits Reasoning in Large Language Models. arXiv preprint arXiv:2201.11903.

Komeili, M., Ribeiro, M. T., Schuster, T., & Zettlemoyer, L. (2021). Internet-Augmented Dialogue Generation. arXiv preprint arXiv:2107.07566.

Lazaridou, A., Razeghi, Y., Athreya, R., & Creswell, A. (2022). Internet-augmented language models through few-shot prompting. arXiv preprint arXiv:2203.05115.

Guu, K., Lee, K., Tung, Z., Pasupat, P., & Chang, M. (2020). Retrieval Augmented Language Model Pre-Training. arXiv preprint arXiv:2004.13922.

Lewis, P., Perez, E., Piktus, A., Petroni, F., Karpukhin, V., Goyal, N., Kenter, T., Moore, A., Lewis, M., & Riedel, S. (2020). Retrieval-Augmented Generation for Knowledge-Intensive NLP Tasks. arXiv preprint arXiv:2005.11401.

Rae, J., Borgeaud, S., Cai, T., Millican, K., Hoffmann, J., Song, H. F., Aslanides, J., Henderson, S., Ring, R., Young, S., Rutherford, E., Hennigan, T., Menick, J., Cassirer, A., Powell, R., van den Driessche, G., Hendricks, L. A., Rauh, M., Huang, P., Glaese, A., Welbl, J., Dyer, C., Summers-Stay, D., Nematzadeh, A., McAleese, N., Shuster, K., Lazaridou, A., Humphreys, M., Goyal, S., Buchatskaya, E., Budden, D., Mourad, S., Such, F. P., Zoph, B., Diver, R., De Las Casas, D., Beattie, C., Sifre, L., Martens, L., Li, X. L., Hassabis, D., Nematzadeh, A., & Irving, G. (2020). Scaling Language Models to 175 Billion Parameters. arXiv preprint arXiv:2007.03879.

Lample, G., Subramanian, S., Smith, E., Denoyer, L., & Ranzato, M. (2019). Multiple Attribute Text Rewriting. arXiv preprint arXiv:1911.09315.

Zhang, T., Wu, F., & Zhuang, Y. (2023). Handling Assumptions and Logical Inconsistencies in Reasoning Systems. arXiv preprint arXiv:2302.03045.

Cheng, Y., Huang, X., & Li, J. (2024). Training Specific Models for Chain-of-Thought Reasoning. Journal of AI Research, 45(3), 789-805.

Bommasani, R., Hudson, D. A., Adeli, E., Altman, R., Arora, S., von Arx, S., Bernstein, M. S., Bohg, J., Bosselut, A., Brunskill, E., Brynjolfsson, E., Buch, V., Cardie, C., Carroll, M., Catanzaro, B., Chatterji, N., Chen, A., Creel, K., Davis, J. Q., Demszky, D., Donahue, C., Doumbouya, M., Durmus, E., Ermon, S., Etchemendy, J., Fei-Fei, L., Finn, C., Gale, T., Gillespie, L., Goel, K., Goodman, N. D.,

Gray, J., Grossman, S., Guha, N., Hashimoto, T. B., Henderson, P., Henke, N., Hewitt, L., Ho, D. E., Hong, J. Y., Jacobs, A. Z., Jurafsky, D., Kalluri, P., Karpathy, A., Keutzer, K., Khani, F., Koh, P. W., Krass, M., Krishnan, R., Ku, M., Landgrebe, T. C. W., Lee, E., Lee, M., Leskovec, J., Levine, S., Lieder, F., Ma, T., Malik, A., Manning, C. D., Mirchandani, K., Mullainathan, S., Narayan, A., Naumann, T., Niebles, J. C., Noseworthy, M., Ong, C. S., Pai, S., Pan, R., Parkes, D., Peters, M. E., Phu, S., Radanovic, G., Raghunathan, A., Reich, R., Rong, Y., Roose, K., Rosenthal, S., Russell, S., Saenko, K., Sagawa, S., Samineni, M., Sattigeri, P., Schain, A., Schmidt, L., Shah, J. A., Shankar, S., Shih, A., Sridhar, D., Steiner, T., Suciu, D., Sundaresan, N., Tambe, M., Tamkin, A., Thomas, D., Trinkwalder, C., Tsvetkov, Y., Tuggener, L., Van Der Vorst, K., Wallach, H., Wang, J., Wang, Y., Webb, R., Whitaker, K., Widdicombe, E., Wilfong, K., Wu, E., Wu, J., Wu, K., Wu, X., Xia, F., Xie, S. M., Yasunaga, M., Yin, D., Zhang, J., Zhang, M., Zhang, X., Zhang, Y., Zhao, A., Zhao, R., Zheng, H., Zheng, L., Zhou, K., Zhou, M., Zou, C., Liang, P., & Manning, C. D. (2021). On the Opportunities and Risks of Foundation Models. arXiv preprint arXiv:2108.07258.

# 7 APPENDIX

## 7.1 - Activity Diagram of our Iterative CoT Solution

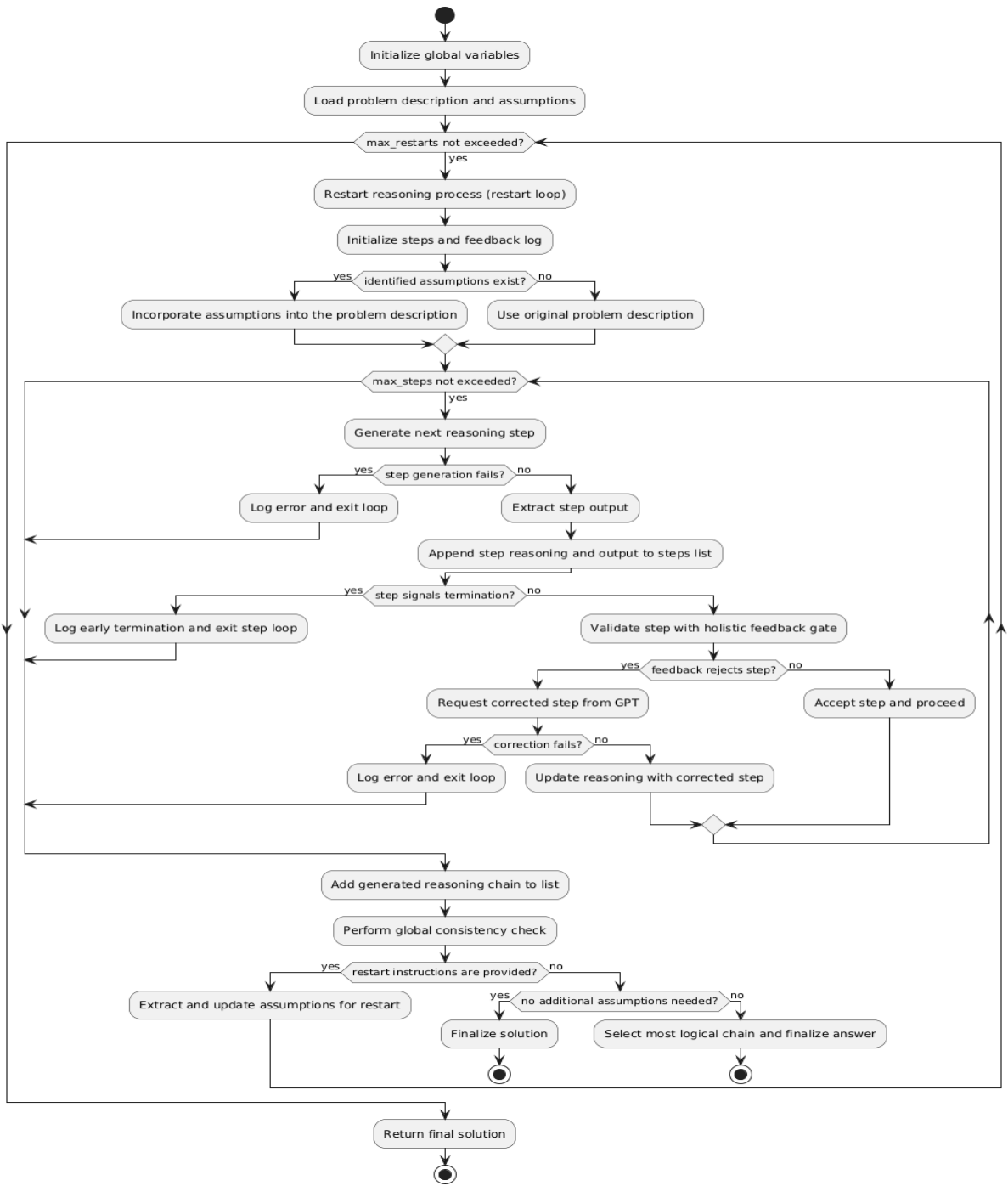

## 7.2 - Results of each specific trial from our study across all models tested-

| GPT-4o-2024-08-06 | | | | | | | |
|---|---|---|---|---|---|---|---|
|  | Trial 1: | Trial 2: | Trial 3: | Trial 4: | Trial 5: | AVG@5 | EAG@5 |
| Question 1: | 0 | 0 | 0 | 0 | 0 | 0% | -25% |
| Question 2: | 0 | 0 | 0 | 0 | 0 | 0% | -25% |
| Question 3: | 1 | 0 | 0 | 0 | 1 | 40% | 30% |
| Question 4: | 0 | 0 | 0 | 0 | 0 | 0% | -25% |
| Question 5: | 1 | 1 | 1 | 0 | 0 | 60% | 45% |
| Question 6: | 0 | 0 | 0 | 0 | 0 | 0% | -25% |
| Question 7: | 0 | 0 | 0 | 0 | 0 | 0% | -25% |
| Question 8: | 1 | 1 | 1 | 1 | 1 | 100% | 150% |
| Question 9: | 0 | 0 | 0 | 0 | 0 | 0% | -25% |
| Question 10: | 0 | 0 | 0 | 0 | 0 | 0% | -25% |

| GPT-4o-2024-08-06 - CoT Enhanced | | | | | | | |
|---|---|---|---|---|---|---|---|
|  | Trial 1: | Trial 2: | Trial 3: | Trial 4: | Trial 5: | AVG@5 | EAG@5 |
| Question 1: | 1 | 1 | 0 | 1 | 1 | 80% | 120% |
| Question 2: | 0 | 0 | 0 | 0 | 0 | 0% | -25% |
| Question 3: | 1 | 1 | 0 | 1 | 1 | 80% | 120% |
| Question 4: | 0 | 0 | 0 | 0 | 0 | 0% | -25% |
| Question 5: | 1 | 1 | 1 | 1 | 0 | 80% | 120% |
| Question 6: | 0 | 0 | 0 | 0 | 0 | 0% | -25% |
| Question 7: | 0 | 0 | 0 | 0 | 0 | 0% | -25% |
| Question 8: | 1 | 1 | 1 | 0 | 1 | 80% | 120% |
| Question 9: | 0 | 0 | 1 | 0 | 1 | 40% | 30% |
| Question 10: | 0 | 0 | 0 | 0 | 0 | 0% | -25% |

| GPT-o1-preview-2024-09-12 | | | | | | | |
|---|---|---|---|---|---|---|---|
|  | Trial 1: | Trial 2: | Trial 3: | Trial 4: | Trial 5: | AVG@5 | EAG@5 |
| Question 1: | 0 | 0 | 0 | 0 | 0 | 0% | -25% |
| Question 2: | 0 | 0 | 0 | 0 | 0 | 0% | -25% |
| Question 3: | 0 | 0 | 0 | 0 | 0 | 0% | -25% |
| Question 4: | 1 | 0 | 0 | 0 | 0 | 20% | 10% |
| Question 5: | 1 | 1 | 1 | 1 | 1 | 100% | 150% |
| Question 6: | 0 | 1 | 1 | 0 | 1 | 60% | 45% |
| Question 7: | 0 | 0 | 0 | 0 | 0 | 0% | -25% |
| Question 8: | 1 | 1 | 1 | 0 | 1 | 80% | 120% |
| Question 9: | 1 | 1 | 1 | 0 | 1 | 80% | 120% |
| Question 10: | 0 | 0 | 0 | 0 | 0 | 0% | -25% |

| Claude-3.5-Sonnet-2024-10-22 | | | | | | | |
|---|---|---|---|---|---|---|---|
|  | Trial 1: | Trial 2: | Trial 3: | Trial 4: | Trial 5: | AVG@5 | EAG@5 |
| Question 1: | 1 | 1 | 1 | 1 | 1 | 100% | 150% |
| Question 2: | 1 | 1 | 1 | 1 | 1 | 100% | 150% |
| Question 3: | 0 | 0 | 1 | 0 | 0 | 20% | 10% |
| Question 4: | 0 | 0 | 0 | 0 | 0 | 0% | -25% |
| Question 5: | 0 | 1 | 1 | 1 | 1 | 80% | 120% |
| Question 6: | 0 | 0 | 0 | 0 | 0 | 0% | -25% |
| Question 7: | 1 | 1 | 1 | 1 | 1 | 100% | 150% |
| Question 8: | 1 | 1 | 1 | 1 | 1 | 100% | 150% |
| Question 9: | 1 | 0 | 0 | 0 | 0 | 20% | 10% |
| Question 10: | 0 | 0 | 0 | 0 | 0 | 0% | -25% |

| Claude-3-Opus-2024-02-29 | | | | | | | |
|---|---|---|---|---|---|---|---|
|  | Trial 1: | Trial 2: | Trial 3: | Trial 4: | Trial 5: | AVG@5 | EAG@5 |
| Question 1: | 1 | 0 | 0 | 0 | 0 | 20% | 10% |
| Question 2: | 0 | 0 | 0 | 0 | 0 | 0% | -25% |
| Question 3: | 0 | 0 | 0 | 0 | 0 | 0% | -25% |
| Question 4: | 0 | 0 | 0 | 0 | 0 | 0% | -25% |
| Question 5: | 1 | 1 | 1 | 1 | 1 | 100% | 150% |
| Question 6: | 0 | 0 | 0 | 0 | 0 | 0% | -25% |
| Question 7: | 0 | 0 | 0 | 0 | 0 | 0% | -25% |
| Question 8: | 1 | 1 | 1 | 1 | 1 | 100% | 150% |
| Question 9: | 0 | 0 | 0 | 0 | 0 | 0% | -25% |
| Question 10: | 0 | 0 | 0 | 0 | 0 | 0% | -25% |

| Claude-3-Opus-2024-02-29- CoT Enhanced | | | | | | | |
|---|---|---|---|---|---|---|---|
|  | Trial 1: | Trial 2: | Trial 3: | Trial 4: | Trial 5: | AVG@5 | EAG@5 |
| Question 1: | 1 | 1 | 1 | 1 | 1 | 100% | 150% |
| Question 2: | 0 | 0 | 0 | 0 | 0 | 0% | -25% |
| Question 3: | 1 | 0 | 1 | 0 | 0 | 40% | 30% |
| Question 4: | 0 | 0 | 0 | 0 | 0 | 0% | -25% |
| Question 5: | 1 | 1 | 1 | 1 | 1 | 100% | 150% |
| Question 6: | 0 | 0 | 0 | 0 | 0 | 0% | -25% |
| Question 7: | 1 | 1 | 1 | 1 | 1 | 100% | 150% |
| Question 8: | 0 | 1 | 1 | 1 | 1 | 80% | 120% |
| Question 9: | 0 | 0 | 0 | 0 | 0 | 0% | -25% |
| Question 10: | 0 | 0 | 0 | 0 | 0 | 0% | -25% |

**7.3 - Total Results across all trials and models used to produce figures-**

| GPT-4o-2024-08-06 | |
|---|---|
| TOTAL AVG | AVERAGE EAG |
| 20% | 5% |
| **GPT-4o-2024-08-06 - CoT Enhanced** | |
| TOTAL AVG | AVERAGE EAG |
| 36% | 39% |
| **GPT-o1-preview** | |
| TOTAL AVG | AVERAGE EAG |
| 34% | 32% |
| **Claude-3.5-Sonnet-20241022** | |
| TOTAL AVG | TOTAL EAG |
| 52% | 67% |
| **Claude-3-Opus-2024-02-29** | |
| TOTAL AVG | TOTAL EAG |
| 22% | 14% |
| **Claude-3-Opus-2024-02-29- CoT Enhanced** | |
| TOTAL AVG | TOTAL EAG |
| 42% | 48% |

**7.4 - GitHub Repository of the solution & csv results file can be found here-**

https://github.com/Soham4001A/LLM_CoT_NoSo_Simple